# Cost-Sharing in Bayesian Knowledge Bases

**Solomon Eyal Shimony** and **Carmel Domshlak**
Dept. of Math. and Comp. Sci.
Ben Gurion University of the Negev
P. O. Box 653, Beer-Sheva 84105, ISRAEL
e-mail: { shimony, carmel} @cs.bgu.ac.il

**Eugene Santos Jr.**
Dept. of Electrical and Comp. Eng.
Air Force Institute of Technology
Wright-Patterson AFB, OH
e-mail:esantos@afit.af.mil

## Abstract

Bayesian knowledge bases (BKBs) are a generalization of Bayes networks and weighted proof graphs (WAODAGs), that allow cycles in the causal graph. Reasoning in BKBs requires finding the most probable inferences consistent with the evidence. The cost-sharing heuristic for finding least-cost explanations in WAODAGs was presented and shown to be effective by Charniak and Husain. However, the cycles in BKBs would make the definition of cost-sharing cyclic as well, if applied directly to BKBs. By treating the defining equations of cost-sharing as a system of equations, one can properly define an admissible cost-sharing heuristic for BKBs. Empirical evaluation shows that cost-sharing improves performance significantly when applied to BKBs.

## 1 INTRODUCTION

Bayes networks [7] are a commonly used reasoning tool within the uncertainty in AI community. Lately, graphical causal probabilistic models have shown up that generalize on the acyclic Bayes networks, in order to cater for causal phenomena which cannot be strictly partially ordered. These models have causal cycles [1, 8], or undirected sections in the directed graphs [2]. Clearly, one still needs to do either belief revision or belief updating [7] in order to perform reasoning in these schemes. These more general models, being less restrictive, pause interesting problems in implementing reasoning algorithms for them.

Bayesian knowledge bases (BKBs) [8] are a generalization of Bayes networks and weighted (AND/OR, directed acyclic) proof graphs (acronym WAODAGs) [4], that allow cycles in the causal graph. Consider the problem of finding the most probable inference ("explanation") consistent with the evidence in a BKB. This problem is analogous to (and more general than) the NP-hard problem of belief revision in Bayes net-works, or finding minimum-cost proof on a WAODAG. As for Bayes networks, reasoning with tree-shaped BKBs can be done efficiently. However, it is clear that in actual applications we cannot usually force our representation to belong to the easy class of problems.

To-date, finding most-probable inference in general BKBs has been implemented as best-first heuristic search, where the heuristic used was cost-so-far, with dismal results. The reason is that this local heuristic does not take into account the cost of nodes (or variables) to be assigned later on in the search. Propagation of costs to be incurred is much preferable, but it is non-trivial to do so in a manner resulting in an admissible heuristic. The latter was first achieved by using the cost-sharing propagated cost method [3] (see next section for a brief definition).

It was shown by Charniak and Husain [3] that for finding least-cost explanations in WAODAGs, the (admissible) cost-sharing heuristic has a much better performance. The cost sharing heuristic was also found useful for belief revision in Bayes networks [9]. Here, we generalize cost-sharing to apply to cyclic graphs, and show that the resulting heuristic is also admissible.

The generalization of the cost-sharing heuristic, while straightforward, causes several problems. First, the cycles in the BKB make the problem of *properly defining* the heuristic nontrivial. If we just used the same defining equations, the fact that there are cycles would make the defining equations cyclic. But by looking at these equations as a *system* of equations, we state that a *solution* to the system is our heuristic. Any such solution to the system of equations is shown to be an admissible heuristic. A second problem is how to solve these equations. The standard top-down algorithm used in prior work would be hindered by the cycles: even if we convert it to a kind of message-passing updating algorithm, in many cases the algorithm will loop indefinitely. Instead, we show that converting the system of semi-linear equations to a linear program, we can evaluate the heuristic in polynomial time.

We begin with a motivating BKB example, followed by a formal definition of BKBs (section 2). We then relate BKBs to WAODAGs, and review the cost shar-ing heuristic for WAODAGs. In Section 4 we extend cost-sharing to handle cycles, and present an efficient method of computing the heuristic. Section 5 discusses several implementation issues and refinements. Section 6 compares search with cost-sharing to search with a local "cost-so-far" heuristic.

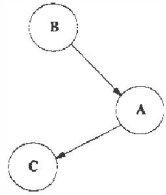

Figure 1: Example graph with $RV$s as nodes.

## 2    BAYESIAN KNOWLEDGE BASES

In modeling an uncertain world, we designate random variables (abbrev. $RV$s) to represent the discrete objects or events in question. We then assign a joint probability distribution to each possible state of the world, i.e., a specific value assignment for each $RV$. Graphical probabilistic models, such as Bayes networks [7], represent the existing dependencies in the model (variables not shown as dependent are assumed independent), facilitating a concise representation of the distribution - as a set of conditional probabilities. Let $D$, $E$ and $F$ be $RV$s. The conditional probability, $P(D|E, F)$, identifies the belief in $D$'s truth given that $E$ and $F$ are both known to be true, and represents an *uncertain causal rule*. We call $D$ the *head* of $P(D|E, F)$ and $\{E, F\}$ the *tail*.

The distribution of the model is defined by,

$$P(A_1, \ldots, A_n) = \prod_{i=1}^{n} P(A_i | X(A_i)) \qquad (1)$$

where $X(A_i)$ is the set of $RV$s which $A_i$ conditionally depends upon. If set $X(A_i)$ is small, the amount of information we must actually store to be able to compute the required joint probability is considerably (exponentially) less than the size of the cross product of the domains.

In Bayes networks, the conditional dependencies are represented with a directed acyclic graph. Let $A$, $B$ and $C$ be $RV$s representing a traffic light, its associated vehicle detector and pedestrian signal, respectively. Suppose that the vehicle detector affects the traffic light, which in turn affects the pedestrian signal. Figure 1 graphically depicts this network over these variables. Since the signal depends upon the light, we say that $A$ is the parent of $C$. Similarly, $B$ is the parent of $A$.

Now, expand the model with an additional variable denoting time of day, and suppose that the domain expert wishes to add the conditional probability that the detector is tripped during rush hour when the light is red. Such an inclusion would introduce a cycle into our graph, which would not be permitted in a Bayes network. In the application domain, however, such cycles are frequently a natural representation.

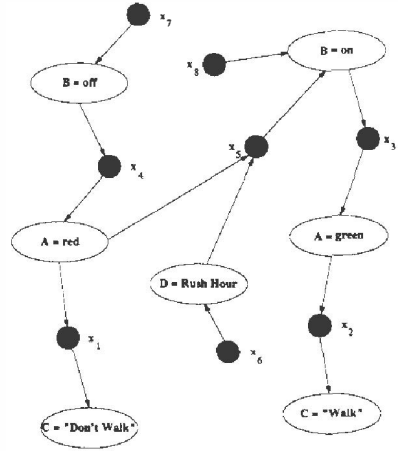

Figure 2: Example Knowledge Graph

By introducing a finer level of distinction than one node per $RV$, using instead one node for each possible $RV$ instantiation, the $BKB$ representation finesses this problem. Assuming the same trio of $RV$s and the partial set of values below:

$$P(C = \text{"Don't Walk"}|A = \text{red}) = x_1$$
$$P(A = \text{green}|B = \text{On}) = x_3$$
$$P(C = \text{"Walk"}|A = \text{green}) = x_2$$
$$P(A = \text{red}|B = \text{Off}) = x_4$$

We can legally add the new constraint, $P(B = \text{On}|A = \text{red}, D = \text{rush hour}) = x_5$, without creating a directed cycle, as shown in Figure 2. Additionally, it is possible to have cycles in the knowledge graph in certain cases, without jeopardizing consistency of the distribution (see [8]).

A $BKB$ graph has two distinct types of nodes. The first, shown as lettered ovals, corresponds to individual $RV$ instantiations. These are called *instantiation nodes* or *I-nodes* for short. The second type of node, depicted as a blackened circle, is called a *support node* or *S-node*. These nodes, which represent the conditional probability value, have exactly one out-bound arrow to the instantiation node representing the head of the conditioning case. Support nodes also have zero or more in-bound dependency or conditioning arrows representing the tail of the conditioning case.

The above representation of the conditional probabilities, by separating out the variable-states and the (possibly partial) conditioning, results both in more flexibility, and a more compact representation [8]. These properties are extremely useful in knowledge acqui-sition and in learning models from data, for various applications such as data-mining [5].

## 2.1 DEFINING KNOWLEDGE GRAPHS

We define the topology as follows:

**Definition 1** *A correlation-graph $G = (I \cup S, E)$ is a directed graph such that $I \cap S = \Phi$ and $E \subseteq \{I \times S\} \cup \{S \times I\}$. Furthermore, for all $a \in S$, $(a, b)$ and $(a, b')$ are in $E$ if and only if $b = b'$. $\{I \cup S\}$ are the nodes of $G$ and $E$ are the edges of $G$. A node in $I$ is called an* instantiation-node *(abbrev. I-node) and a node in $S$ is called a* support-node *(abbrev. S-node).*

I-nodes represent the various states of the world such as the truth or falsity of a proposition. S-nodes, on the other hand, explicitly embody the relationships between the I-nodes.

Let $\pi$ be a partition on $I$. Intuitively, $\pi$ denotes the groups of I-nodes (states) which are mutually exclusive. This can be used to represent random variables with discrete but multiple instantiations, with each partition cell corresponding to an *RV*.

**Definition 2** *$G$ is said to* I-respect *$\pi$ if for all cells $\sigma$ in $\pi$, for any S-node $b \in S$ such that $(b, a) \in E$, $b$ does not have a parent in $\sigma$ except, possibly, $a$.*

Basically, mutually exclusive I-nodes cannot be directly related to each other through the S-nodes. Next, we define mutual exclusion between S-nodes.

**Definition 3** *Two S-nodes $b_1$ and $b_2$ in $S$ are said to be* mutually exclusive *with respect to $\pi$ if there exist different I-nodes $c_1, c_2$ that are parents of $b_1, b_2$, respectively and $c_1, c_2$ are in the same cell in $\pi$.*

**Definition 4** *$G$ is said to* S-respect *$\pi$ if for all I-nodes $a$ in $I$, any two distinct parents of $a$ (S-nodes $b_1$ and $b_2$) are mutually exclusive.*

**Definition 5** *$G$ is said to* respect *$\pi$ if $G$ both I-respects and S-respects $\pi$.*

To complete our knowledge-graph, we define a function $w$ from $S$ to $\Re$. This serves as the mechanism for handling uncertainty in the relationships.

**Definition 6** *A knowledge-graph $K$ is a 3-tuple $(G, w, \pi)$ where $G = (I \cup S, E)$ is a correlation-graph, $w$ is a function from $S$ to the positive reals (for each $a \in S$, $w(a)$ is the weight of $a$), $\pi$ is a partition on $I$, and $G$ respects $\pi$.*

The probabilistic semantics of a knowledge graph is provided by relating weights to probabilities, as follows: $P'(a) = e^{-w(a)}$, where $P'(a)$ is the conditional probability that the child of $a$ is true given that the parents of $a$ are true. To make sure that the probabilities obey the axioms of probability theory, a normalization constraint is enforced [8] on BKBs. However, this issue is irrelevant to finding most-probable inference, and is thus beyond the scope of this paper.

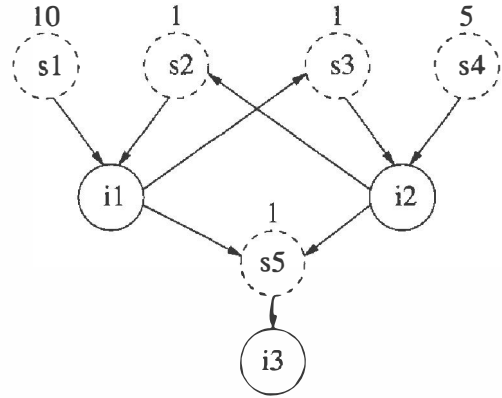

Figure 3: Knowledge Graph with a Cycle

## 2.2 INFERENCE GRAPHS

A BKB is a knowledge graph, together with an inference scheme. The latter is defined by a set of permissible inference graphs. An inference graph is a subgraph of the knowledge graph corresponding to an inference chain (or proof). Let $r = (I' \cup S', E')$ be some subgraph of our correlation-graph $G = (I \cup S, E)$ where $I' \subseteq I$, $S' \subseteq S$, and $E' \subseteq E$. Furthermore, $r$ has a *weight* $\omega(r)$ defined as follows:

$$\omega(r) = \sum_{s \in S'} w(s).$$

An I-node $a \in I'$ is said to be *well-supported* in $r$ if it has an incoming S-node in $r$ (that is, if there exists an edge $(b, a)$ in $E'$). An S-node $b$ is said to be *well-founded* in $r$ if all its incoming I-nodes (conditions) are also present in $r$. An S-node $b \in S'$ is said to be *well-defined* in $r$ if it supports some I-node.

$r$ is said to be an *inference* over $K$ if it is acyclic, consistent (i.e. for all cells $\sigma$ in $\pi$, $|I' \cap \sigma| \leq 1$), all of its I-nodes are well supported, and all of its S-nodes are well-founded and well-defined. An inference thus corresponds to a proof. Given the knowledge graph in Figure 3, one possible inference can be seen in Figure 4.

For an abductive BKB, the problem we are addressing is, given a set $s$ of I-nodes, find an inference $r$ of minimum weight that contains all of $s$. Given the semantics of costs in BKBs, such an inference (proof) is equivalent to a maximum probability explanation (abductive inference) for $s$.

## 3   COST-SHARING IN WAODAGS

WAODAGs [4] are essentially acyclic knowledge graphs, with a single sink (out-degree 0) node $s$ (called the evidence node), and a partition of nodes into AND nodes (corresponding to S-nodes in a BKB) and OR nodes (I-nodes in a BKB). The evidence node is an AND node. Each WAODAG node has an associated cost (or weight). A proof $r$ of $s$ is a subgraph containing $s$ where for each AND node in $r$, all of its parents are in $r$, and for each OR node in $r$, at least one of its parents are in $r$. Proofs in WAODAGs correspond to inferences in BKBs. The partition function $\pi$ has no counterpart in WAODAGs.

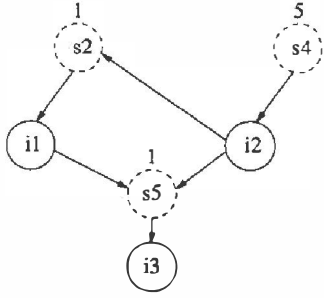

Figure 4: An Inference in the Knowledge Graph

The cost of a proof $r$ is the sum of the weights of all nodes in $r$. As for knowledge graphs, one would like to find a least cost proof that contains the evidence node $s$. This is an NP-hard problem, usually solved by best-first search, starting from $s$, and adding parents when necessary (branching when several possibilities exist - at OR nodes). An obvious admissible heuristic, cost-so-far, estimates the cost of a partial proof $p$ as the sum of costs of nodes currently in $p$.

The heuristic can be improved upon by propagating costs, but preserving admissibility is non-trivial. Such an improved heuristic, cost-sharing, was first presented in [3], where the search is in terms of edges, rather than nodes. We review the cost-sharing heuristic for WAODAGs below, beginning with some necessary notation borrowed from [3].

Let edge $e = (a, b)$ (from node $a$ to node $b$); we call $a$ the source of $e$, and $b$ the sink of $e$. Also, we say that $b$ is a child (immediate descendent) of $a$. If $e$ is an edge, then $v_e$ denotes that node $v$ is its source, and $u^e$ denotes that $u$ is its sink (we also sometimes use the notation $v_e$ also to denote the source node of $e$, likewise for sinks). If $v$ is a node, then $e_v$ is an arbitrary edge incoming to $v$, and $e^v$ an outgoing edge from $v$. Also, $E_v$ is the set of all incoming edges of $v$, and $E^v$ is the set of all outgoing edges. A node with no parents is called a root node.

For convenience, a dummy edge $e'$ leading from the evidence node is added to the graph, as well as one dummy edge leading into each root node. A state in the search space is a cut of the WAODAG (a set-wise minimal set of edges that separates $s$ from the root nodes). The initial state is the set $\{e'\}$, and a final state is a cut consisting only of root node dummy edges. The actual solution is the set of nodes abutting the dummy edges. Since we need to find the minimal cost solution, we need a heuristic value for each cut. In fact, the heuristic value is defined over both edges and nodes, as follows. Let $w(v)$ be the weights of the root nodes. We define the heuristic cost function $c$ from $E \cup V \cup 2^E$ to the non-negative reals as:

$$c(v) = \begin{cases} w(v) & \text{if v is a root node} \\ c(E_v) & \text{if v is an AND node} \\ \min_{e_v \in E_v} c(e_v) & \text{if v is an OR node} \end{cases}$$

for nodes,

$$c(e^v) = \frac{c(v_e)}{|E^v|}$$

for edges, and for sets of edges $C \subseteq E$:

$$c(C) = \sum_{e \in C} c(e)$$

## 4    COST-SHARING IN BKBS

For BKBs, we intend to use the same definitions for the cost-sharing heuristic. One difference between BKBs and WAODAGs is that in WAODAGs, only root nodes have weights, whereas in BKBs every S-node has a weight. The difference can be overcome by observing that for each S-node $v$ we can always add one new I-node and S-node pair (call the latter $v'$), set $w(v') = w(v)$, and let the new $w(v)$ be 0. The semantics of the BKB stay the same, and now only root nodes have non-0 cost. Instead of doing that, we will note that the new I-node only has one parent and one child, and absorb $w(v')$ into the equation for $v$, to get:

$$c(v) = \begin{cases} c(E_v) + w(v) & \text{if v is an S node} \\ \min_{e_v \in E_v} c(e_v) & \text{if v is an I node} \end{cases} \quad (2)$$

for nodes, and the same equations as above for edges. Noting, however, the optimization in [3], observe that disjoint S-nodes are never in the same inference, and thus we can replace the equation for edges by:

$$c(e^v) = \frac{c(v_e)}{k(e^v)} \quad (3)$$

where $k(e^v)$ is the number of consistent immediate support paths. Specifically, if $v$ is an S-node, then $k(e^v) = 1$ since there is only one outgoing edge from an S-node. If $v$ is an I-node, $k(e^v)$ is the number of consistent I-nodes immediately supported by S-nodes that are children of $v$.[1] As for WAODAGs, we have for sets of edges:

$$c(C) = \sum_{e \in C} c(e) \quad (4)$$

---

[1] This number should be an upper bound on the number of edges outgoing from $v$ that are in any inference. It may be possible to get a tighter bound in some cases, and if so that bound can be used in place of $k(e^v)$.

It is by no means clear whether these equations are sufficient to uniquely define the cost function. However, treating equations 2, 3, 4 as a system of equations in the variables $c(v), c(e)$ with $v \in V, e \in E$ rather than a definition, we can refer to *solutions* of the system. Henceforth, we will denote by $c$ an arbitrary solution to equations 2, 3, 4, whenever unambiguous. We will show that an arbitrary solution $c$ to the system (henceforth called a cost-sharing solution) is an admissible heuristic, by extending the proof of [3].

**Theorem 1** *Any cost-sharing solution $c$ is an admissible heuristic for BKBs.*

Proof outline: we first note that while the BKB is a graph with cycles, an inference is acyclic, and corresponds to an AND-DAG. A cut of an inference is defined exactly as for an AND-DAG in [3]: a minimal set of edges that any path from the evidence to the leaves must intersect. As in WAODAGs, we define a cut of the BKB as a cut of some BKB inference.

Now, we proceed with the same proof as in [3]. All steps of the proof are the same, it does not matter that we have cycles, as the cycles only serve to further decrease $c$, and thus it is still an underestimate of the true weight.

The remaining problems are with applying the WAODAG expansion operator $S_\tau$, which requires a topological sort $\tau$ of the DAG. Since we have cycles, this is no longer possible. We must guarantee that once the we apply the expansion operator at a node, its outgoing edges will not be used anymore. To do that, we modify the expansion operator as follows. A state $s$ is a set of edges and a set of deleted edges. Our expansion operator $S$, applied at node $n$ is the same as $S_\tau$, except that when $S$ is applied at node $n$, the edges $E^n$ are added to the set of deleted edges. A state which contains a deleted edge is illegal, and discarded.

In order that all possible inferences be reachable, it is not sufficient to apply the expansion according to a topological ordering. If $S$ is applied, at each state, at all nodes where there is some $e^n$ in the current set of edges, reachability is maintained. $\square$

We now address the problem of computing a solution $c(v)$. Clearly, the seemingly obvious solution of using the equations directly will not help: some values will be undefined initially. One *could* think of a scheme that, to compute a minimum over several terms, some defined and some not, just takes the minimum over the currently defined terms, and propagates the resulting value. If every node participates in some inference, such a scheme is guaranteed to assign a cost value at each node eventually. However, in order to have all equations satisfied, it may be necessary to update cost values already derived, for example, due to finding a lower value than already used before, at an I-node. It turns out that such a scheme may loop indefinitely, as the following example shows.

Consider the knowledge graph in figure 3. The following values can be computed immediately: $c((s1, i1)) = c(s1) = 10$, and $c((s4, i2)) = c(s4) = 5$. All other equations now contain undefined terms, so we proceed by evaluating partially defined minima. For example, we could now set (temporarily), $c(i2) = 5$. As a result, we get the edge cost $c((i2, s2)) = 2.5$, since $i2$ has two children. We now have $c(s2) = 3.5$, and this in turn makes $c(i1) = 3.5$, $c((i1, s3)) = 1.75$, and $c(s3) = 2.75$. This causes a re-evaluation of $c(i2) = 2.75$, which causes re-evaluation to proceed indefinitely, until eventually (in the limit of an infinite number of loops, or in practice determined by computational numerical accuracy) we get convergence at: $c(i1) = c(i2) = c(s2) = c(s3) = 2$, $c((i1, s2)) = c((i1, s3)) = 1$, and $c(i3) = c((s5, i3)) = c(s5) = 3$. Note further that the costs we get reflect an illegal, cyclic inference, but since we need an *underestimate* in order to get an admissible heuristic, this is not a problem.

Solving the system of equations efficiently is nontrivial. In fact, if we had max functions in addition to the min functions, or if the summation included negative terms, it would be easy to show that the problem of finding a solution is NP-hard (and deciding the existence of a solution is NP-complete). However, in our case, we can use linear programming techniques to derive a solution, by transforming the equations to a linear system, as follows. For each node and edge, we have a variable, which for convenience we denote by the same name. Linear equations are left as they are. Minimization equations are translated as follows:

$$v = min(u_1, u_2, ..., u_k)$$

is replaced by the set of inequalities:

$$v \le u_1, \quad v \le u_2, \quad ... \quad v \le u_k$$

Observe that the latter set of inequalities is weaker than the minimization. Finally, the objective function to maximize is:

$$\Theta(c) = \sum_{v \in I} c(v)$$

An optimal solution $c^*$ to the above linear program can be found using standard linear programming methods, such as the simplex method [6]. A solution always exists, since setting all $c(v) = 0$, $v \in I$ clearly determines a unique, not necessarily optimal, solution to the equations and inequalities.

**Theorem 2** *Let $c^*$ be an optimal solution to the linear program. Then $c^*$ is also a solution to the cost-sharing equation system (equations 2, 3, 4).*

Proof: Let $c^*$ be an optimal solution to the linear program. Assume that $c^*$ violates some of equations 2, 3, 4. Since the linear linear program equations are the same as equations 2, 3, 4, except that minimization was replaced by inequalities, only equations of the form:

$$c(v) = \min_{e_v \in E_v} c(e_v)$$

can be violated. Let $v$ be a variable (node) where the above equation is violated. Then, since the linear program enforces $c^*(v) \leq c^*(e_v)$ for all $e_v \in E_v$, then it must be the case that for this variable, $c^*(v) < c^*(e_v)$ for all $e_v \in E_v$ (otherwise it will indeed be the minimum, thus not violating the equation). Define $R$ to be the set of nodes consisting of $v$ and its immediate descendents (all immediate descendents are S-nodes), and let $E^R$ be all edges with sources in $R$. Define another solution $c^{*'}$ to the linear program as follows. For every node $u$ not in $R$ let $c^{*'}(u) = c^*(u)$, and likewise for every edge $e$ not in $E^R$, let $c^{*'}(e) = c^*(e)$. Let $c^{*'}(v) = v_{min} = \min_{e_v \in E_v} c^*(e_v)$, and let the linear program equations (starting with edges $v_E$, then the S-nodes, then the rest of the edges in $E^R$) determine the as yet undefined costs in $c^{*'}$. The resulting solution is unique, because it uses equations for which all values on the right-hand-side are already determined, and the left-hand side is not. Clearly, $c^{*'}$ is also a solution to the linear program[2], where only one I-node cost was changed (increased). We have $\Theta(c^{*'}) > \Theta(c^*)$, and thus $c^*$ is not optimal, a contradiction. $\square$

As an example, consider the graph of Figure 3, where we would get the following set of inequalities:

$$i1 \leq 10 \;\; ; \;\;\; i1 \leq 1 + (i2, s2)$$
$$i2 \leq 5 \;\; ; \;\;\; i2 \leq 1 + (i1, s3)$$
$$(i1, s3) = \frac{i1}{2} \;\; ; \;\;\; (i2, s2) = \frac{i2}{2}$$
$$s5 = (i1, s5) + (i2, s5) \;\; ; \;\;\; i3 = s5$$
$$(i2, s5) = \frac{i2}{2} \;\; ; \;\;\; (i1, s5) = \frac{i1}{2}$$

where we need to maximize $c(i1) + c(i2) + c(i3)$. The optimal solution is the same as the convergence value shown above, i.e. $c(i1) = c(i2) = 2$ and $c(i3) = 3$.

## 5    IMPLEMENTATION DETAILS

In applying the cost-sharing heuristic, we actually use a simplified linear program, as follows. First, whenever an I-node $v$ has only a single parent, we use $c(v) = c(e_v)$ rather than the inequality. We also cancel out all edge cost variables by substituting them according to equations (3, 4). Finally, we can also cancel out by substitution all the S-node costs (noting that S-nodes all have only one child), to get a system of inequalities just for the I-node costs. For I-nodes with more than one parent, we get:

$$c(v^e) \leq w(u_e) + \sum_{e'^w \in E_u} \frac{c(w_{e'})}{k(e'^w)} \qquad (5)$$

---

[2]Costs of edges and S-nodes can only be increased by this change, thus can (at worst) affect variables not in $R$ through introducing higher values on the right-hand-side of the inequalities for other I-nodes, which cannot cause the inequalities to be violated.

and, for I-nodes with just one parent:

$$c(v^e) = w(u_e) + \sum_{e'^w \in E_u} \frac{c(w_{e'})}{k(e'^w)} \qquad (6)$$

Next, observe that the linear program is only necessary within each strongly connected component. The implementation is, thus:

1. Initialization: find strongly connected components, and sort them in a total ordering consistent with a topological ordering of the components, such that the evidence node(s) is first.

2. Add to the graph a dummy edge $(*, v)$ for every S-node $v$.

3. Proceed from the last component down to the first, and for each component do:

   (a) Set up the linear program over variables determined by nodes and edges for the current component (including edges to and from other components). Incoming edges will have costs set in previous components, if any, and these costs are considered as constants for this component.

   (b) Solve the linear program, to get the heuristic costs.

4. Initialize an agenda with a single state $s$, with edges$(s)=e'$, the evidence dummy edge, and an empty list of expanded nodes.

5. Looping until time limit, or required number of solutions found, get state $s$ of lowest heuristic cost from agenda, and do:

   (a) Find the first strongly connected component containing a node $v$ with some outgoing edge $e^v$ in edges$(s)$.

   (b) If there is no such component, output $s$ (a solution).

   (c) Otherwise, expand $s$ at the current strongly connected component, as follows. For each unexpanded node $v$ in the current component for which there is an edge $d^v \in E^v$ in edges$(s)$, and for each edge $e_v \in E_v$ such that no parent of $u_e$ (the source node of $e_v$) has been expanded do:

      i. create a new state $s'$, with node $v$ added to the list of expanded nodes. The edges of $s'$ are the edges of $s$ with all edges $E^v$ removed and all edges $E_u$ added.

      ii. Evaluate the cost of $s'$ by subtracting the cost of removed edges and adding the cost of added edges, from the cost of $s$.

      iii. If $s'$ is consistent (does not contain any pair of I-nodes from the same cell in the partition), insert it into the agenda.

For example, let us trace the algorithm as run on the BKB fragment of figure 3, with $i3$ being the evidence.

The heuristic costs are computed as shown in the previous section. The strongly connected components are $\{i1, i2\}$ and $\{i3\}$. The starting state, $S_0$, contains just the dummy edge $(i3, *)$. Search proceeds as shown in table 1, where "Pop" is the ID of the popped state. For lack of space, the dummy edges $(*, i5), (*, s2), (*, s3)$ are missing from the table, but this should not adversely affect clarity.

In the actual implementation, several details should be observed. First, instead of maintaining a list of deleted edges, it is sufficient, and more efficient, to treat any edge outgoing from an expanded node as if it were a deleted edge, and maintain a list of expanded I-nodes. Second, the fact that S-nodes are all AND nodes with a single outgoing edge is used to save some time: once an edge outgoing from an S-node is picked, we are forced to select all the incoming edges into the S-node anyway, so we do all that in one expansion step, and do not keep track of expanded S-nodes.

## 6   EXPERIMENTAL RESULTS

The above algorithm was tested on several BKB's produced from an available acyclic BKB for reasoning about raising gold-fish. The network has 165 I-nodes and 350 S-nodes. Cycles were introduced by random reversal of several arc pairs. Evidence selection was also random. Runtime and number of expansion steps (iterations through step 4 of the algorithm) were compared between a search algorithm using cost-sharing and one using cost-so-far. The program was implemented in C++, and run on a SPARC-10.

Results are depicted in Figures 5, 6 using log-scale of time to solution and number of expansion steps, respectively for the Y axis (X axis is just the number of the problem instance, and thus essentially meaningless here). Times for cost-sharing include initialization of the heuristic costs. The cases labeled cost-so-far (failed) are those taking 2000 seconds without reaching a solution, or crashing due to lack of swap space.

Figure 7 plots total number of problems solved vs. total CPU time. Cost-sharing does better than cost-so-far by at least one order of magnitude. Finding several best solutions is also useful [9]. A timing comparison for the 10 best solutions is depicted in Figure 8.

These preliminary experiments suggest that cost-sharing is an extremely useful search heuristic for graphs with cycles, as well as directed acyclic graphs. We know of no other heuristics for this search problem. Nor is it clear how one would apply schemes such as clustering to BKBs, and even if they could, a strongly-connected component size of over 40 for most of the problem instances in the experiments suggest that the clique size would be too large to handle by such schemes. Thus, heuristic search with cost-sharing appears to be the only viable method for BKBs not in one of the (topologically) easier classes of problems.

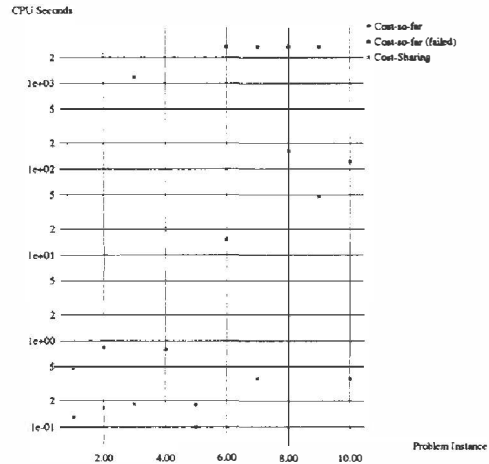

Figure 5: Time: Cost-Sharing vs. Cost-So-Far

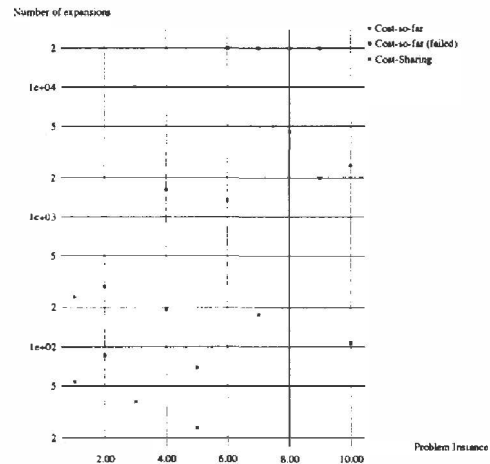

Figure 6: Expansion Count Comparison

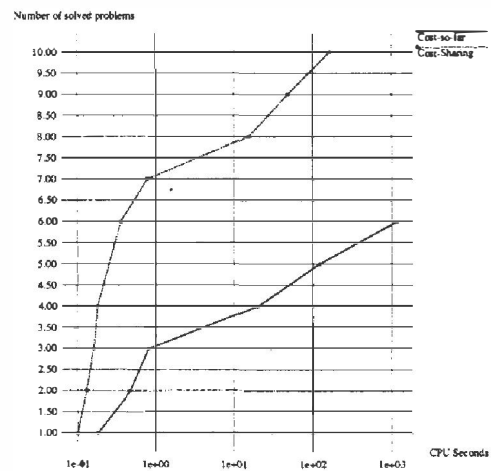

Figure 7: Number of Problems Solved vs. CPU Time

| Iteration | Pop | Edges | Expand Edge | Delete Edges | Add Edges | New State ID | Cost |
|---|---|---|---|---|---|---|---|
| 0 | | | | | $(i3,*)$ | $S_0$ | 3 |
| 1 | $S_0$ | $(i3,*)$ | $(i3,*)$ | $(i3,*)$ | $(i1,s5),(i2,s5)$ | $S_1$ | 3 |
| 2 | $S_1$ | $(i1,s5),(i2,s5)$ | $(i1,s5)$ | $(i1,s5)(i1,s3)$ | $(*,s1)$ | $S_2$ | 12 |
| | | | $(i1,s5)$ | $(i1,s5),(i1,s3)$ | $(i2,s2)$ | $S_3$ | 4 |
| | | | $(i2,s5)$ | $(i2,s2),(i2,s5)$ | $(*,s4)$ | $S_4$ | 7 |
| | | | $(i2,s5)$ | $(i2,s5),(i2,s2)$ | $(i1,s3)$ | $S_5$ | 4 |
| 3 | $S_5$ | $(i1,s3),(i1,s5)$ | $(i1,s5)$ | $(i1,s5),(i1,s3)$ | $(*,s1)$ | $S_6$ | 12 |
| 4 | $S_3$ | $(i2,s2),(i2,s5)$ | $(i2,s5)$ | $(i2,s5),(i2,s2)$ | $(*,s4)$ | $S_7$ | 7 |
| 5 | $S_7$ | $(*,s4)$ | NONE | | | | 7 |

Table 1: Trace of the Search Algorithm

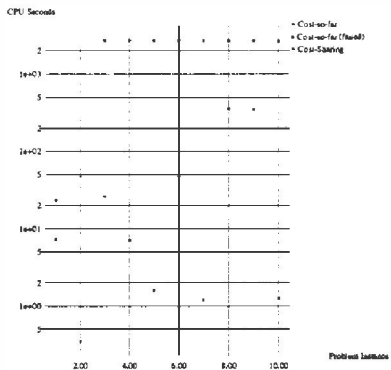

Figure 8: Time Comparison for 10 Best Solution

# 7  CONCLUSION

Bayesian knowledge bases (BKBs) are a useful generalization of both weighted proof graphs and Bayes networks that allow causal cycles. The cycles in the graph pose a difficult problem for implementing reasoning in the model. A heuristic search algorithm currently appears the only viable method to perform belief revision (finding best "inference"). We adapted a successful WAODAG search heuristic, the cost-sharing heuristic [3], to apply to BKBs. The adaptation is non-trivial due to the fact that the heuristic as-is, is undefined for cyclic graphs. Additionally, it is non-trivial to compute the modified heuristic efficiently.

Having successfully modified the cost-sharing heuristic to BKBs, we showed empirically that the heuristic saves considerable search effort in several BKBs which is a toy version of an application domains: raising goldfish. In addition to using the suggested algorithm and heuristic for BKB reasoning henceforth, it may be possible to use the scheme for reasoning in other models with cycles, such as reasoning in chain graphs or the model presented in [1].

### Acknowledgments

This research is supported in part by an infrastructure grant for data-mining from the Israeli Ministry of Science, and by AFOSR Project #940006.